\documentclass[conference]{IEEEtran}

\usepackage{cite}
\usepackage{graphicx}
\usepackage{amsmath,amssymb,amsfonts}
\usepackage{algorithmic}
\usepackage{textcomp}
\usepackage{xcolor}
\usepackage{booktabs}
\usepackage{multirow}
\usepackage{array}
\usepackage{tabularx}
\usepackage{float}
\usepackage{url}
\usepackage{hyperref}
\usepackage{balance}
\usepackage[scaled]{helvet}
\usepackage{caption}

\hypersetup{
    colorlinks=true,
    linkcolor=blue,
    citecolor=blue,
    urlcolor=blue
}

\begin{document}


\title{A Configurable Privacy-Preserving MRI Processing Workflow Using Deep Learning-Based Brain Extraction and Adaptive Anatomical Preservation}


\author{

\IEEEauthorblockN{Rayeef Ali Khan}
\IEEEauthorblockA{
School of Computing (Artificial Intelligence)\\
Dublin City University\\
Dublin, Ireland\\
rayeef.alikhan2@mail.dcu.ie
}

\and

\IEEEauthorblockN{Komal Raj Mahantesh}
\IEEEauthorblockA{
School of Computing (Data Analytics)\\
Dublin City University\\
Dublin, Ireland\\
komalraj.mahantesh2@mail.dcu.ie
}
}
\maketitle
\begin{abstract}

Structural Magnetic Resonance Imaging (MRI) is widely used in neuroimaging research and clinical practice due to its ability to provide detailed visualisation of brain anatomy [1], [3]. The increasing availability of publicly accessible neuroimaging datasets has accelerated the development of artificial intelligence (AI)-based medical image analysis techniques while simultaneously introducing significant challenges related to patient privacy [1], [2], [23]. Existing deep learning-based brain extraction methods generally generate a single fixed preprocessing output, providing limited flexibility for balancing privacy protection with the preservation of anatomically relevant structures [3], [6], [7].

This paper presents a configurable privacy-preserving MRI processing workflow that extends deep learning-based brain extraction through adaptive anatomical preservation, interactive preservation selection, and systematic quality control procedures. The proposed methodology employs SynthStrip for automated brain extraction [3], followed by morphological mask expansion to generate configurable shell-based preservation levels that allow researchers to select preservation strategies according to different analytical and privacy requirements. The workflow further incorporates an Interactive Preservation Framework for user-guided preservation selection and an automated Quality Control Framework that provides multi-plane visualisation and brain-mask overlay verification to improve transparency and reproducibility throughout the preprocessing pipeline.

The proposed workflow was implemented in Python using
open-source neuroimaging libraries within the Renku
reproducible research environment and evaluated using
structural T1-weighted MRI data from the publicly available IXI
dataset. Experimental results demonstrate that the workflow
consistently generates anatomically plausible brain extraction
and configurable preservation outputs while providing systematic
visual verification of processing quality.

The principal contribution of this work is a modular and reproducible MRI preprocessing framework that enhances conventional deep learning-based brain extraction with configurable anatomical preservation, interactive user-guided processing, and integrated quality control mechanisms. The workflow provides a practical foundation for privacy-preserving neuroimaging research and establishes a platform for  developments in secure medical image processing and collaborative AI applications.

\end{abstract}




\section{Introduction}

Structural Magnetic Resonance Imaging (MRI) has become an indispensable imaging modality for neuroscience research and clinical practice due to its ability to provide high-resolution visualisation of brain anatomy [1], [3]. Advances in artificial intelligence (AI) and deep learning have further accelerated the use of structural MRI for automated brain segmentation, disease diagnosis, image-guided analysis, and computer-aided clinical decision-making [4]–[7]. Simultaneously, the increasing availability of publicly accessible neuroimaging datasets has promoted collaborative research and the development of increasingly sophisticated machine learning models [24], [25].

Despite these advances, sharing structural MRI data presents significant privacy challenges [1], [2], [23]. Unlike conventional medical images, structural MRI volumes often retain facial and cranial anatomical information that may allow patient re-identification if appropriate anonymisation procedures are not applied [1], [2]. Consequently, privacy-preserving preprocessing has become an essential stage within modern neuroimaging workflows, particularly for multi-institutional collaborations, public data repositories, and artificial intelligence research involving sensitive medical imaging data [1], [23].

Brain extraction (skull stripping) is a widely adopted preprocessing technique for removing non-brain tissue before downstream neuroimaging analysis [14], [12], [16]. Recent deep learning-based approaches, including SynthStrip, have significantly improved the robustness and generalisability of automated brain extraction across heterogeneous MRI datasets [3]. However, these methods are primarily designed to generate a single fixed preprocessing output [3], [6], [7]. Although suitable for many applications, they provide limited flexibility when studies require different balances between privacy and anatomical preservation.

To address this limitation, this paper proposes a configurable privacy-preserving MRI processing workflow that extends deep learning-based brain extraction through adaptive anatomical preservation, interactive preservation selection, and systematic quality-control procedures. Rather than replacing existing segmentation algorithms, the proposed methodology builds upon SynthStrip [3] by introducing a modular post-processing framework capable of generating multiple preservation levels from a single brain extraction result. Researchers can select preservation configurations that best satisfy study-specific privacy and analytical requirements while maintaining a reproducible workflow.
The workflow was implemented using Python and open-source neuroimaging libraries within the Renku reproducible research environment [30] and evaluated using structural T1-weighted MRI volumes from the publicly available IXI dataset [27]. The workflow integrates automated brain extraction, adaptive morphological preservation, interactive user-guided preservation selection, and multi-plane quality-control visualisation into a unified processing pipeline designed to improve transparency, flexibility, and reproducibility during MRI preprocessing.

The main contributions of this paper are summarised as follows:

\begin{itemize}
    \item Development of a configurable privacy-preserving MRI preprocessing workflow that integrates deep learning-based brain extraction, adaptive anatomical preservation, interactive preservation selection, and systematic quality-control procedures within a unified and reproducible framework.

    \item Design of an Adaptive Anatomical Preservation Framework that generates configurable shell-based preservation levels from a single brain segmentation, enabling flexible trade-offs between patient privacy and anatomical context while remaining independent of the underlying brain extraction algorithm.

    \item Implementation of an Interactive Preservation Framework that enables researchers to visually compare multiple shell-based preservation configurations and select the most appropriate output according to specific privacy requirements and downstream analytical objectives.

    \item Development of a systematic Quality Control Framework incorporating multi-plane visualisation and brain-mask overlay verification to support transparent inspection of brain extraction accuracy and preservation quality.

    \item Construction of a modular and reproducible implementation using Python, SynthStrip, open-source neuroimaging libraries, and the Renku computational environment, providing a practical foundation for future extensions in privacy-preserving neuroimaging research.
\end{itemize}


\section{Related Work}

Structural Magnetic Resonance Imaging (MRI) preprocessing is a fundamental stage in neuroimaging analysis, providing the foundation for brain segmentation, tissue classification, disease diagnosis, image registration, and numerous artificial intelligence (AI)-based medical imaging applications [12], [13], [16]. Accurate brain extraction is particularly important because the presence of non-brain tissues, including the skull, scalp, and facial structures, can negatively influence downstream analytical tasks and introduce variability in automated image analysis pipelines. Consequently, brain extraction has remained an active area of research for more than two decades [14], [12].

Traditional neuroimaging preprocessing frameworks, including the Brain Extraction Tool (BET) [14], FreeSurfer [12], Statistical Parametric Mapping (SPM) [15], and FSL [16], rely on deformable surface models, intensity-based segmentation, atlas-guided techniques, and registration-based frameworks such as Advanced Normalization Tools (ANTs) [11] to separate brain tissue from surrounding anatomical structures [13]. Affine registration techniques also remain fundamental components of many neuroimaging preprocessing workflows for aligning brain images across acquisitions and subjects [18]. Although these methods have been widely adopted in neuroimaging research, their performance may vary across different scanner manufacturers, imaging protocols, and patient populations.

More recently, deep learning-based approaches have demonstrated improved robustness by learning complex anatomical representations directly from large-scale neuroimaging datasets [4]–[7]. Among these methods, SynthStrip has shown strong generalisability across multiple MRI contrasts and acquisition conditions while requiring minimal dataset-specific parameter tuning [3].

Despite the considerable progress achieved in automated brain extraction, existing methods primarily focus on accurate segmentation and generally produce a single fixed preprocessing output [3], [6], [7]. While this behaviour is appropriate for many neuroimaging applications, it provides limited flexibility when different studies require different balances between patient privacy and anatomical preservation. In practice, some applications benefit from aggressive removal of peripheral anatomical structures to maximise anonymisation, whereas others require preservation of boundary regions that may contain anatomically relevant information for subsequent processing.

Several MRI anonymisation techniques have been proposed to address privacy concerns by removing or obscuring identifiable facial features prior to data sharing [1], [2], [23]. These approaches reduce the risk of patient re-identification; however, they often operate independently of brain extraction workflows and provide limited opportunities for configurable preservation of surrounding anatomical structures. Also, existing preprocessing pipelines rarely integrate systematic quality-control procedures that allow researchers to verify preservation behaviour consistently across processed MRI volumes.

Motivated by these limitations, this paper proposes a configurable privacy-preserving MRI preprocessing workflow that extends deep learning-based brain extraction through adaptive anatomical preservation, interactive preservation selection, and systematic quality-control procedures. Rather than introducing a new segmentation network, the proposed methodology builds upon the robustness of SynthStrip while providing researchers with greater flexibility, reproducibility, and transparency during MRI preprocessing [3].

\begin{table}[!t]
\caption{Comparison of Representative Brain Extraction Approaches}
\label{tab:comparison}
\centering
\footnotesize
\renewcommand{\arraystretch}{1.2}
\setlength{\tabcolsep}{3pt}

\begin{tabular}{lccccc}
\toprule
\textbf{Method} & \textbf{DL} & \textbf{FO} & \textbf{AP} & \textbf{IPS} & \textbf{IQC} \\
\midrule
BET         & $\times$ & \checkmark & $\times$ & $\times$ & L \\
FreeSurfer  & $\times$ & \checkmark & $\times$ & $\times$ & L \\
SynthStrip  & \checkmark & \checkmark & $\times$ & $\times$ & L \\
Proposed    & \checkmark & \checkmark & \checkmark & \checkmark & \checkmark \\
\bottomrule
\end{tabular}

\vspace{1mm}
\footnotesize
DL: Deep Learning; FO: Fixed Output; AP: Adaptive Preservation;\\
IPS: Interactive Preservation Selection; IQC: Integrated Quality Control; L: Limited.
\end{table}

Table~\ref{tab:comparison} compares representative brain extraction approaches with the workflow. Conventional methods primarily focus on automated brain extraction and generate a single preprocessing output. In contrast, the workflow extends deep learning-based brain extraction by incorporating configurable anatomical preservation, interactive preservation selection, and integrated quality-control procedures within a unified and reproducible MRI preprocessing framework.


\section{Proposed Methodology}

\subsection{Overview of the Proposed Workflow}

The workflow is a modular privacy-preserving MRI preprocessing workflow extending deep learning-based brain extraction with adaptive preservation, interactive preservation, and systematic quality control. Instead of developing a new brain extraction algorithm, the workflow builds on SynthStrip and introduces configurable post-processing for balancing privacy and anatomical preservation across neuroimaging applications [3].

\begin{figure}[!t]
\centering
\includegraphics[width=\columnwidth]{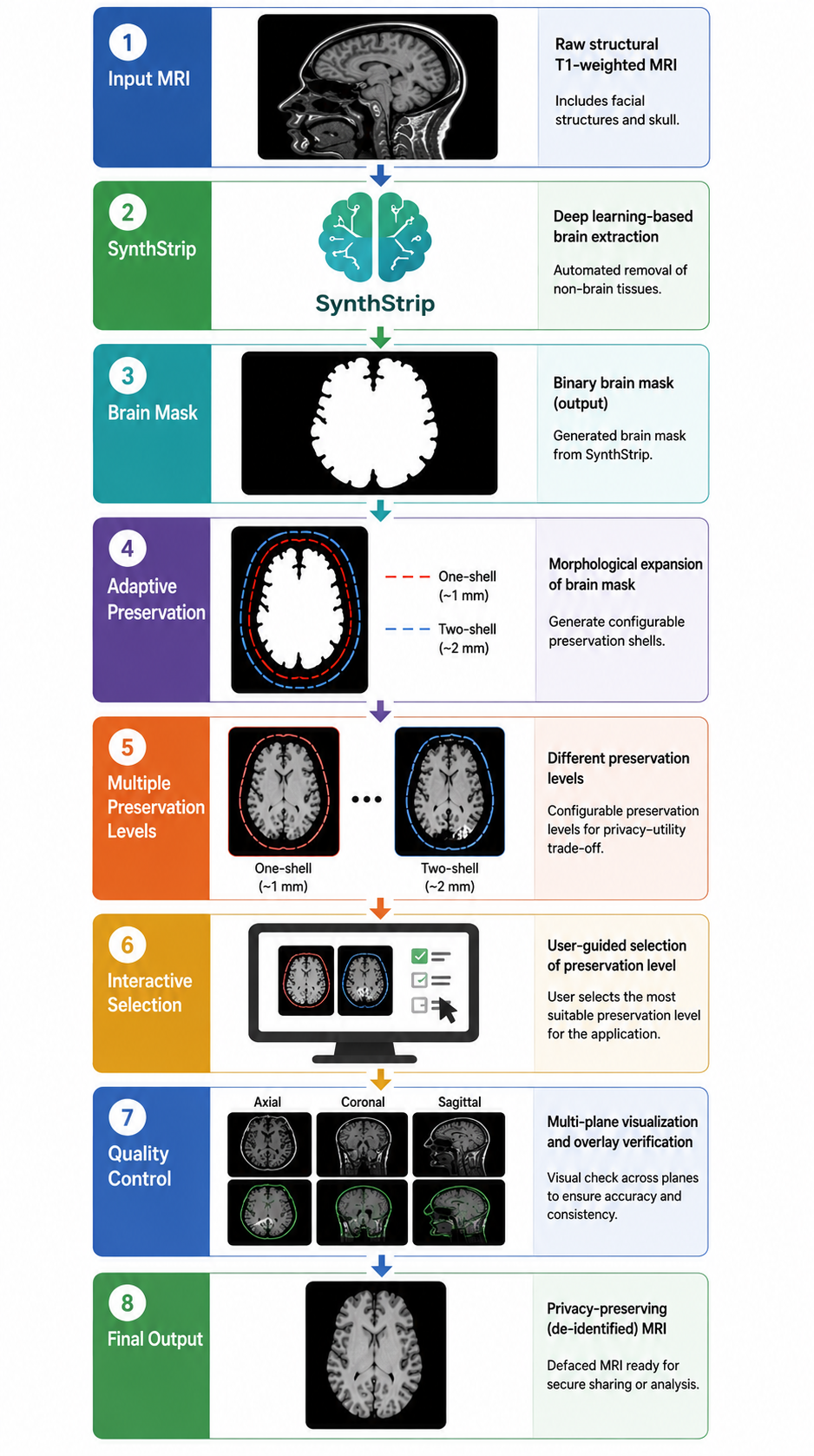}
\caption{Overview of the proposed privacy-preserving MRI processing workflow. The pipeline begins with a raw structural T1-weighted MRI containing facial structures, followed by deep learning-based brain extraction using SynthStrip, shell-based adaptive preservation, interactive preservation selection, integrated quality control, and generation of a privacy-preserving brain-extracted MRI suitable for secure sharing and downstream neuroimaging analysis.}
\label{fig:workflow}
\end{figure}

As illustrated in Fig.~\ref{fig:workflow}, the workflow consists of five sequential processing stages. First, structural T1-weighted MRI volumes are provided as input to SynthStrip, which automatically generates a brain mask and corresponding brain-extracted image [3]. Second, the extracted brain mask undergoes adaptive morphological expansion to generate multiple preservation margins surrounding the segmented brain. Third, the generated preservation configurations are presented through an Interactive Preservation Framework, allowing comparison of preservation levels before selecting the preferred configuration. Fourth, an automated Quality Control Framework produces multi-plane visualisations and brain-mask overlay images that facilitate systematic verification of processing quality. Finally, the selected preservation output is exported as a privacy-preserving MRI volume suitable for subsequent neuroimaging analysis or collaborative data sharing.

The modular architecture keeps processing stages independent while remaining compatible with the workflow. Future brain extraction or preservation improvements can be incorporated without modifying the workflow, improving reproducibility, maintainability, and extensibility.

\subsection{Deep Learning-Based Brain Extraction}

Brain extraction is the foundation of the workflow, separating intracranial brain tissue from surrounding anatomy. Accurate extraction is essential because adaptive preservation operates directly on the brain mask.

The workflow uses SynthStrip as the baseline brain extraction method because of its robust performance across diverse MRI protocols [3]. Unlike traditional skull-stripping techniques that often require parameter tuning or are sensitive to scanner-specific characteristics, SynthStrip utilises a deep learning-based segmentation model to generate anatomically consistent brain masks with minimal user intervention [3]. Rather than developing a new segmentation network, this research leverages the reliability of SynthStrip as a preprocessing component and focuses on extending its functionality through configurable post-processing and adaptive preservation.

For each structural T1-weighted MRI volume, SynthStrip generates both a binary brain mask and a brain-extracted image. The binary mask defines the spatial boundary of the segmented brain and serves as the primary input to the Adaptive Anatomical Preservation Framework. By separating brain extraction from subsequent preservation operations, the workflow maintains a modular architecture that enables alternative segmentation algorithms to be incorporated in future implementations without modifying the preservation or quality-control stages.

This design choice enhances reproducibility and long-term extensibility while ensuring that the workflow remains compatible with future advances in deep learning-based brain extraction. Consequently, the contribution of this work lies not in replacing existing segmentation methods but in extending their practical utility through configurable anatomical preservation and integrated workflow management.

\begin{figure}[!tb]
\centering
\includegraphics[width=\columnwidth]{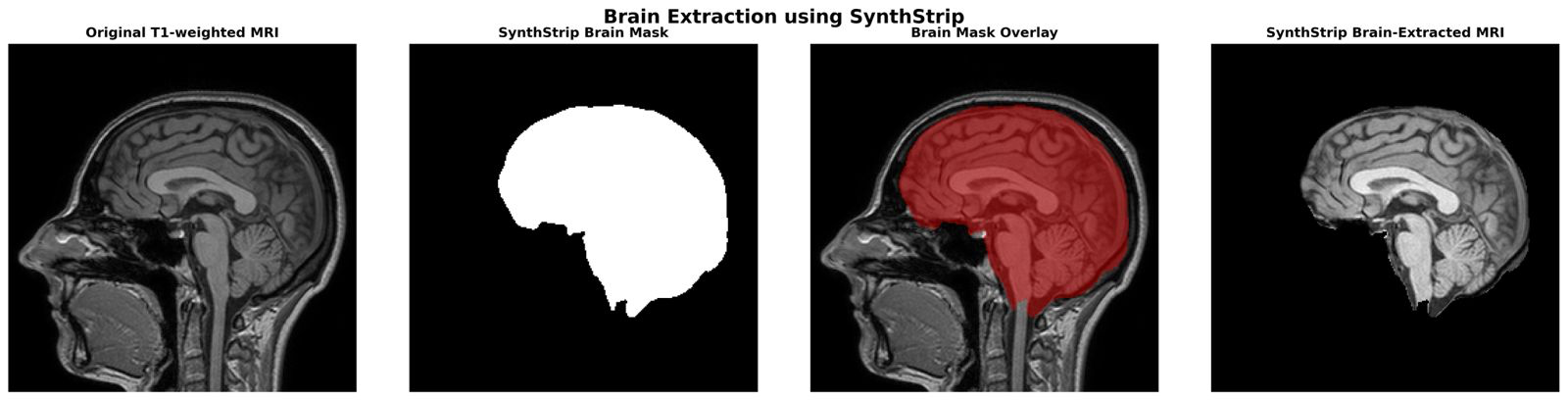}
\caption{Deep learning-based brain extraction using SynthStrip. (a) Original structural T1-weighted MRI, (b) generated binary brain mask, (c) brain mask overlaid on the original MRI for segmentation verification, and (d) resulting brain-extracted MRI. The extracted brain mask serves as the input to the proposed shell-based adaptive preservation framework.}
\label{fig:brain_extraction}
\end{figure}

\subsection{Adaptive Anatomical Preservation}

Conventional brain extraction methods are primarily designed to generate a single brain-extracted output by removing non-brain tissues surrounding the intracranial region. While this approach is appropriate for many neuroimaging applications, it offers limited flexibility when different studies require varying levels of anatomical context or privacy protection. Research involving collaborative data sharing, algorithm development, or image preprocessing may benefit from preserving selected boundary regions while still reducing the risk of patient identification. These differing requirements motivated the development of the proposed Adaptive Anatomical Preservation Framework.

The proposed framework operates as a post-processing stage following deep learning-based brain extraction. Rather than modifying the underlying segmentation algorithm, the framework utilises the binary brain mask generated by SynthStrip as the foundation for configurable preservation. Morphological expansion is applied to the extracted brain mask to generate shell-based preservation regions surrounding the original brain boundary [22]. Each expanded mask represents a different level of anatomical preservation while maintaining the integrity of the segmented brain.

The use of morphological expansion offers two important advantages. First, preservation behaviour becomes independent of the underlying segmentation model, allowing the framework to remain compatible with future brain extraction algorithms without requiring architectural modifications. Second, multiple preservation configurations can be generated from a single brain extraction result, eliminating the need to repeat computationally expensive segmentation procedures whenever different preservation levels are required.

\begin{figure}[!tb]
\centering
\includegraphics[width=\columnwidth]{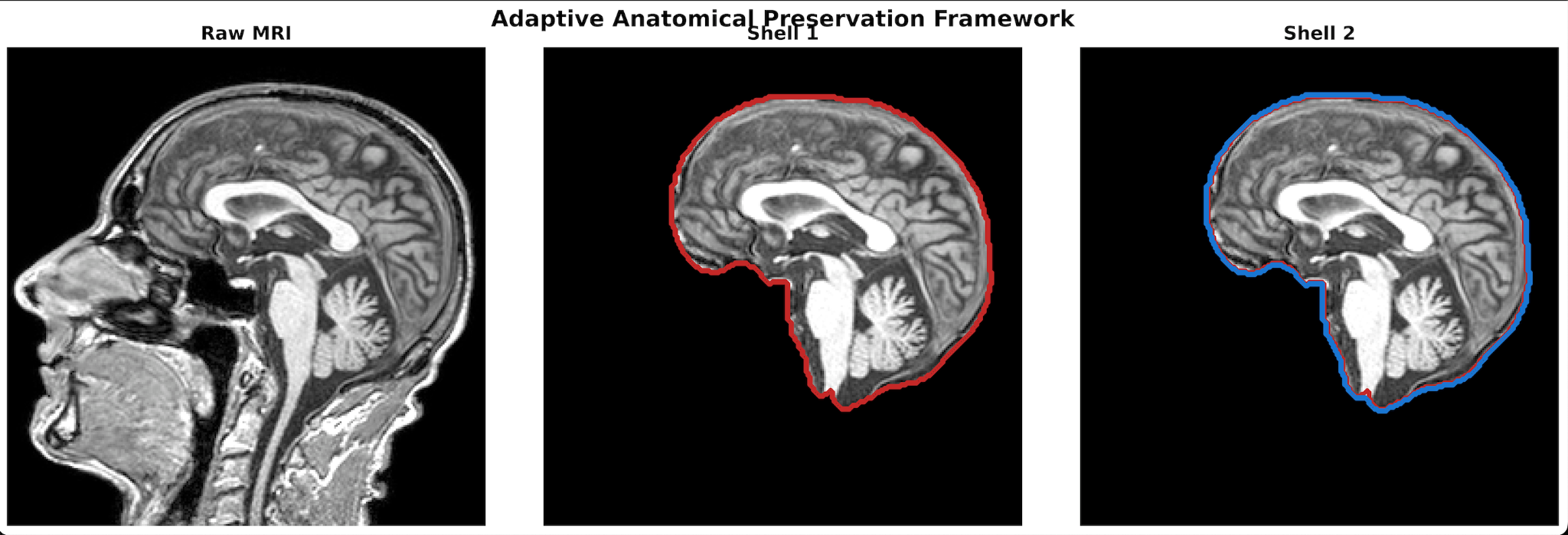}
\caption{Shell-based adaptive anatomical preservation generated through morphological expansion of the SynthStrip brain mask. The one-shell (approximately 1 mm) configuration preserves a minimal anatomical boundary surrounding the extracted brain, while the two-shell (approximately 2 mm) configuration retains additional peripheral anatomical context. The coloured outlines highlight the incremental preservation achieved through successive shell expansion.}
\label{fig:adaptive_preservation}
\end{figure}

In this study, the adaptive preservation framework generated configurable shell-based preservation levels by performing morphological expansion of the SynthStrip-generated brain mask. Two preservation configurations were evaluated, corresponding to one-shell (approximately 1 mm) and two-shell (approximately 2 mm) expansions beyond the original brain boundary. As shown in Fig.~\ref{fig:adaptive_preservation}, the one-shell configuration prioritises stronger anonymisation by preserving only a minimal boundary surrounding the extracted brain, whereas the two-shell configuration retains additional anatomical context that may be beneficial for downstream neuroimaging analysis.

Rather than enforcing a single preservation strategy, the proposed framework enables researchers to select the preservation level that best satisfies the privacy requirements and analytical objectives of a particular study. Because the preservation process is implemented independently of the segmentation stage, the proposed methodology enhances the practical utility of existing deep learning-based brain extraction techniques without altering their internal operation. This modular design enables configurable anatomical preservation to be incorporated into established neuroimaging preprocessing pipelines while maintaining reproducibility, extensibility, and compatibility with future developments in medical image segmentation.

To further support adaptive preservation, the proposed workflow incorporates a lightweight rule-based recommendation framework for shell selection. Rather than relying solely on visual inspection, each preservation configuration is quantitatively evaluated using objective preservation metrics. Candidate shell configurations are recommended when they satisfy predefined heuristic criteria, providing a transparent and reproducible decision-support mechanism while allowing researchers to override the recommendation according to the objectives of a particular neuroimaging study [22].

\begin{itemize}
    \item Boundary Confidence $>$ 5\%
    \item Relative Volume Gain between 5\% and 15\%
    \item Boundary Distance $<$ 5 voxels [22].
    \item Estimated Privacy Exposure below the predefined threshold [23].
\end{itemize}

A shell satisfying these criteria is recommended as the preferred preservation level, providing an objective balance between anatomical preservation and patient privacy while maintaining flexibility for different downstream neuroimaging applications [23].

\subsection{Interactive Preservation Framework}

Conventional MRI preprocessing workflows generally generate a single output, requiring researchers to either accept the default preprocessing result or repeat the complete processing pipeline using different parameter settings. Such approaches can be inefficient when multiple preservation strategies must be evaluated for different neuroimaging applications. To address this limitation, the proposed workflow first applies a rule-based recommendation framework during adaptive preservation to identify suitable shell configurations based on quantitative preservation metrics. The recommended configuration, with all generated preservation levels, is then presented through the Interactive Preservation Framework (IPF), enabling researchers to review, accept, or override the recommendation without requiring repeated brain extraction.

Following adaptive anatomical preservation, the workflow automatically generates multiple shell-based preservation outputs corresponding to different preservation levels. These outputs are presented simultaneously, allowing researchers to visually compare the effects of each preservation level on anatomical boundary retention and surrounding tissue removal. This comparative approach supports informed decision-making by enabling users to evaluate the trade-off between patient privacy and anatomical preservation before selecting a final preprocessing configuration.

Unlike conventional fixed-output preprocessing pipelines, the proposed Interactive Preservation Framework recognises that different neuroimaging studies may have different preprocessing requirements. Research focused on maximum anonymisation may favour smaller preservation margins, whereas studies involving anatomical measurements or advanced image analysis may benefit from retaining additional peripheral structures. Rather than assuming a universally optimal preservation level, the framework provides configurable preservation options that can be selected according to specific research objectives while maintaining a consistent and reproducible processing workflow.

\begin{figure}[!tb]
\centering
\includegraphics[width=\columnwidth]{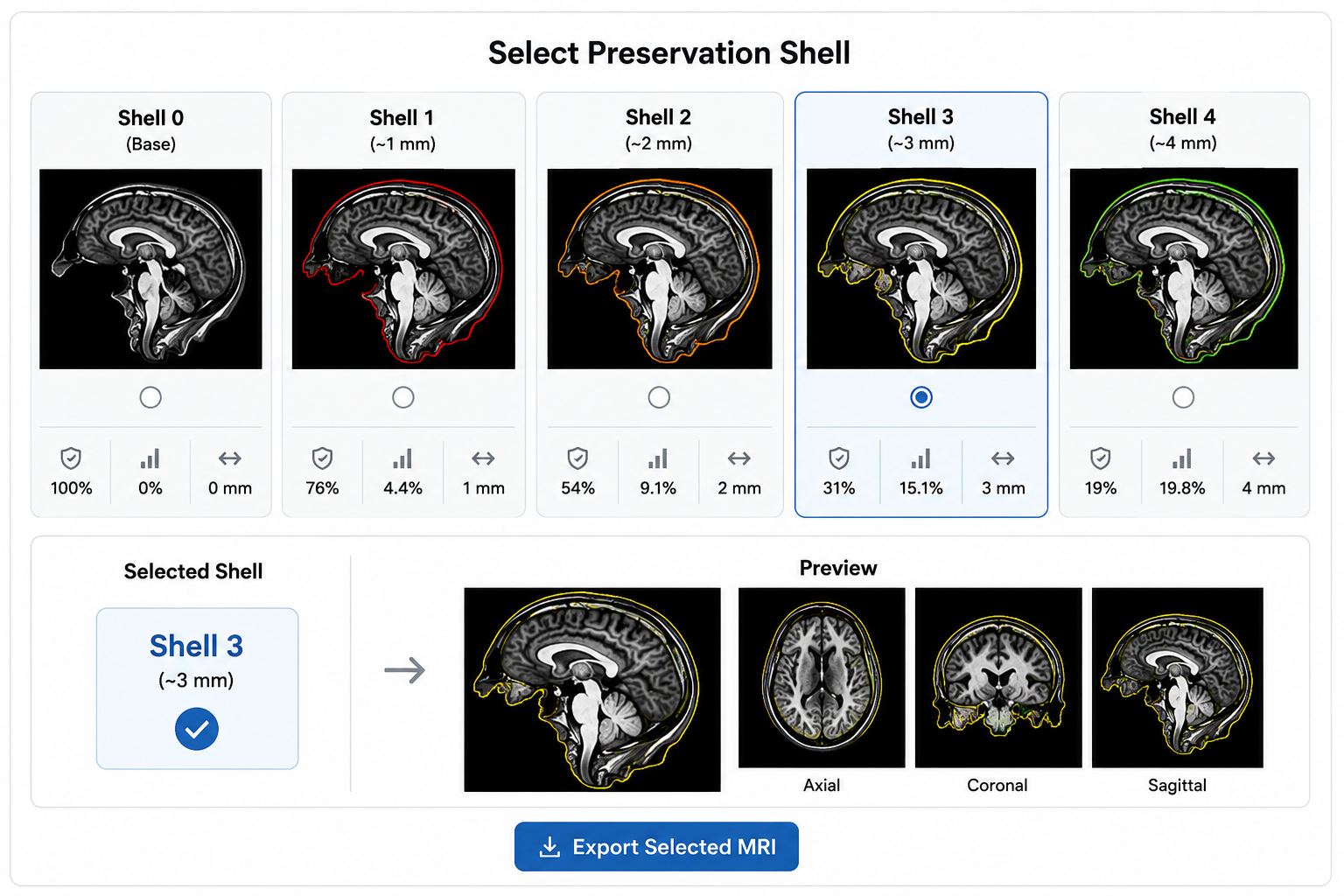}
\caption{Interactive Preservation Framework enabling comparison of multiple preservation configurations generated from a single brain extraction result. Researchers can visually evaluate preservation levels and select the configuration that best satisfies the privacy and analytical requirements of a particular neuroimaging study.}
\label{fig:interactive_preservation}
\end{figure}

As illustrated in Fig.~\ref{fig:interactive_preservation}, researchers can visually compare multiple preservation configurations generated from a single brain extraction result before selecting the most appropriate configuration for downstream neuroimaging analysis.

Another advantage of the Interactive Preservation Framework is improved computational efficiency. Since all preservation configurations are generated from a single brain extraction result, researchers can evaluate multiple preservation strategies without repeatedly executing the segmentation model. This reduces processing time while improving flexibility and transparent preprocessing.

By incorporating user-guided preservation selection into the overall preprocessing pipeline, the proposed methodology extends conventional automated brain extraction workflows with configurable decision support. This improves usability while reinforcing flexibility, reproducibility, and privacy-aware preprocessing.

\subsection{Quality Control Framework}

Reliable verification is essential for reproducible neuroimaging workflows. Although modern deep learning-based brain extraction methods provide highly accurate segmentation results, preprocessing errors may still occur because of variations in image quality, acquisition protocols, or anatomical characteristics. Such errors can propagate into downstream analyses if they remain undetected. To address this challenge, the workflow incorporates an integrated Quality Control Framework (QCF) that enables systematic visual verification of brain extraction and adaptive preservation results.

The Quality Control Framework automatically generates representative visualisations. Multi-plane inspection is performed by displaying representative axial, coronal, and sagittal slices, allowing researchers to evaluate preservation behaviour throughout the complete three-dimensional image rather than relying on a single anatomical view. This comprehensive visual assessment improves confidence that anatomically relevant brain structures have been preserved while unnecessary peripheral tissues have been appropriately removed.

In addition to multi-plane visualisation, the workflow generates brain mask overlay images in which the segmentation boundaries are superimposed on the original structural MRI volume. These overlays verify spatial agreement between the extracted brain mask and underlying anatomy, allowing researchers to identify potential segmentation inaccuracies, boundary artefacts, or excessive tissue removal before subsequent image analysis.

Unlike conventional preprocessing pipelines, such as fMRIPrep [17], that often rely on external tools for quality assessment, the proposed framework integrates verification directly within the processing workflow. This design improves transparency by ensuring that every preservation configuration can be inspected immediately after generation using a consistent and reproducible quality-control procedure. Consequently, preprocessing decisions become more interpretable while reducing the likelihood of undetected processing errors.

\begin{figure}[!tb]
\centering
\includegraphics[width=\columnwidth]{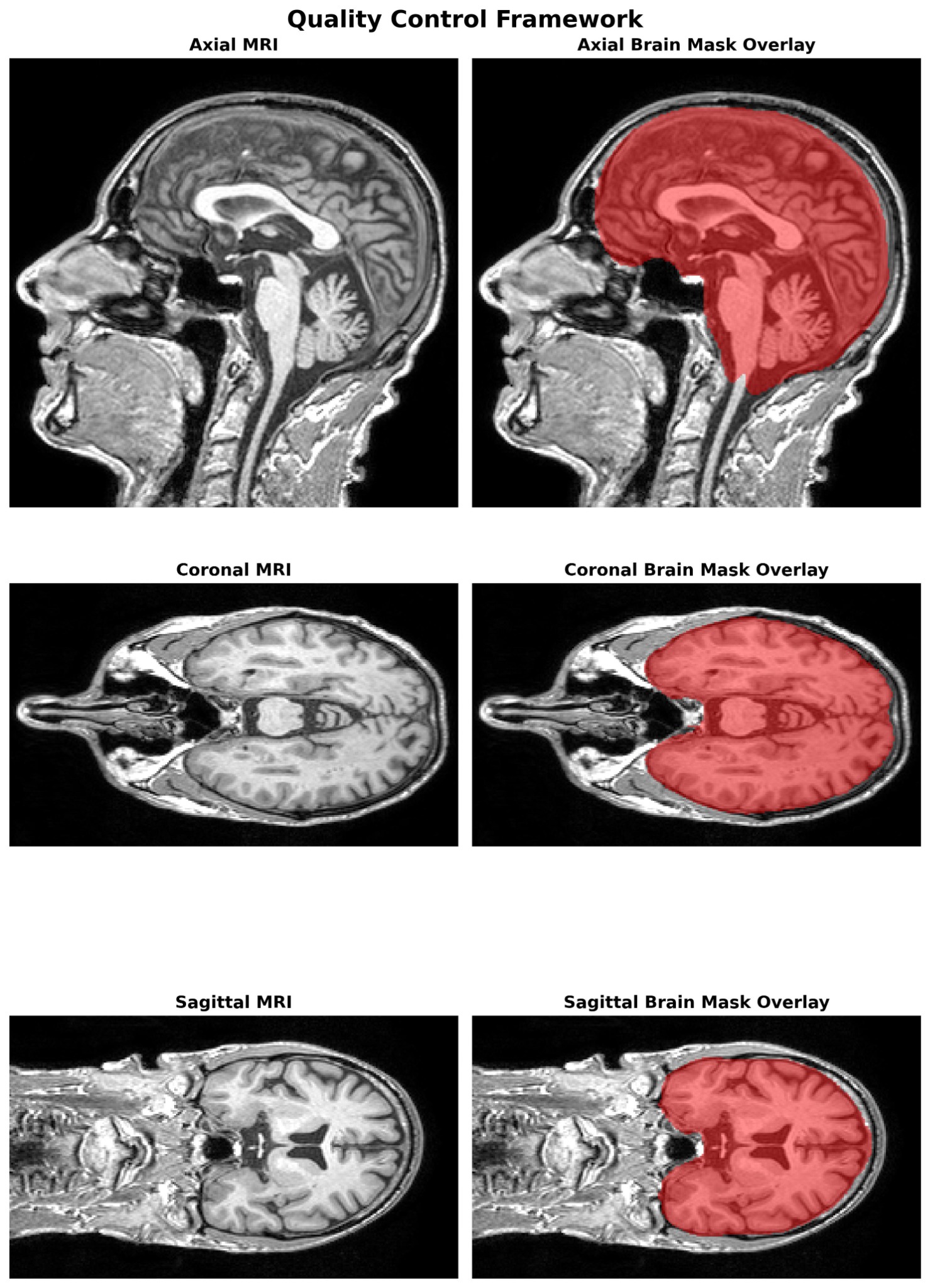}
\caption{Quality Control Framework showing representative axial, coronal, and sagittal visualisations with brain mask overlay verification. Integrated quality-control procedures enable systematic inspection of brain extraction accuracy and preservation behaviour across processed structural MRI volumes.}
\label{fig:quality_control}
\end{figure}

As illustrated in Fig.~\ref{fig:quality_control}, the integrated Quality Control Framework provides representative multi-plane visualisations with brain mask overlays, enabling systematic verification of brain extraction accuracy and preservation behaviour across processed structural MRI volumes.

By combining automated brain extraction, adaptive anatomical preservation, interactive preservation selection, and integrated quality-control procedures within a unified workflow, the proposed methodology provides a reproducible preprocessing framework that supports both privacy preservation and reliable preparation of structural MRI data for downstream neuroimaging research.

The proposed methodology combines deep learning-based brain extraction, adaptive anatomical preservation, interactive preservation selection, and integrated quality-control procedures within a modular and reproducible preprocessing framework. By separating brain extraction from configurable preservation and verification stages, the workflow provides greater flexibility than conventional fixed-output preprocessing pipelines while remaining compatible with future segmentation algorithms and privacy-enhancing technologies.


\section{Experimental Setup}

\subsection{Dataset}

The proposed workflow was evaluated using structural T1-weighted Magnetic Resonance Imaging (MRI) volumes obtained from the publicly available Information eXtraction from Images (IXI) dataset [27]. The IXI dataset contains high-quality brain MRI scans acquired from healthy subjects using multiple MRI scanners and acquisition protocols [27], making it suitable for evaluating reproducible neuroimaging preprocessing workflows. In this study, structural T1-weighted MRI volumes were selected because they represent the standard imaging modality for brain extraction and structural neuroimaging analysis.

Prior to processing, all MRI volumes were verified for compatibility with the workflow and organised within a reproducible project structure. No additional preprocessing operations, such as N4 bias field correction [19], were applied to evaluate the workflow using consistent input data. Each volume served as an independent input to the processing pipeline, enabling consistent evaluation of brain extraction, adaptive anatomical preservation, interactive preservation selection, and quality-control procedures across representative subjects.

The evaluation was conducted using a total of 581 T1-weighted MRI volumes from the IXI dataset, comprising 464 training subjects (80\%) and 117 validation subjects (20\%). Segmentation performance was assessed on the validation set using the Dice Similarity Coefficient (DSC), Intersection over Union (IoU), Precision, Recall, and F1 Score.

\subsection{Implementation Details}

The proposed workflow was implemented in Python using open-source neuroimaging and scientific computing libraries. Automated brain extraction was performed using SynthStrip, while image loading, processing, and visualisation were implemented using libraries including NiBabel, NumPy, SciPy, and Matplotlib. The modular implementation separated brain extraction, preservation, interactive processing, and quality-control procedures into independent components, facilitating reproducibility and future extensibility.

Experimental execution was performed within the Renku reproducible research environment [30], ensuring a consistent computational configuration throughout all experiments. Renku provided version-controlled execution and a structured environment for organising datasets, processing scripts, generated preservation outputs, and quality-control visualisations. This reproducible implementation supports transparent experimentation and enables future researchers to replicate or extend the workflow under comparable conditions.

\begin{table}[!t]
\caption{Implementation Details of the Proposed Workflow}
\label{tab:implementation}
\centering
\renewcommand{\arraystretch}{1.2}
\begin{tabular}{ll}
\toprule
\textbf{Component} & \textbf{Specification} \\
\midrule
Dataset & IXI Dataset \\
MRI Modality & Structural T1-weighted MRI \\
Brain Extraction & SynthStrip \\
Programming Language & Python \\
Image Processing & NiBabel, NumPy, SciPy \\
Visualisation & Matplotlib \\
Execution Environment & Renku \\
Workflow Type & Modular Privacy-Preserving MRI Processing \\
\bottomrule
\end{tabular}
\end{table}


\section{Results and Discussion}

\subsection{Quantitative Evaluation}

To objectively evaluate the performance of the deep learning-based brain extraction stage, quantitative segmentation was performed on the validation subset of the IXI dataset, comprising 117 previously unseen T1-weighted MRI volumes. The remaining 464 MRI volumes were used during model development. Segmentation performance was assessed using standard medical image segmentation metrics [21], [22], including the Dice Similarity Coefficient (DSC), Intersection over Union (IoU), Precision, Recall, and F1 Score.

The proposed brain extraction model achieved a Dice Similarity Coefficient of 98.73\%, an Intersection over Union of 97.48\%, a Precision of 99.05\%, a Recall of 98.41\%, and an F1 Score of 98.73\%, as summarised in Table~\ref{tab:metrics}. These results indicate excellent agreement between the predicted brain masks and the reference annotations, demonstrating accurate delineation of brain tissue with minimal false-positive and false-negative predictions.

\begin{table}[!t]
\caption{Quantitative Performance of Brain Extraction}
\label{tab:metrics}
\centering
\renewcommand{\arraystretch}{1.2}
\begin{tabular}{lc}
\toprule
\textbf{Metric} & \textbf{Value (\%)} \\
\midrule
Dice Similarity Coefficient (DSC) & 98.73 \\
Intersection over Union (IoU) & 97.48 \\
Precision & 99.05 \\
Recall & 98.41 \\
F1 Score & 98.73 \\
\bottomrule
\end{tabular}
\end{table}

The high Dice Similarity Coefficient and IoU demonstrate excellent overlap between the predicted brain masks and the reference annotations, while the high Precision and Recall indicate accurate preservation of brain tissue with minimal inclusion of surrounding non-brain structures.

The consistently high Precision and Recall values indicate that the segmentation model effectively preserves brain tissue while minimising the inclusion of surrounding non-brain structures. Since the proposed Adaptive Anatomical Preservation Framework operates directly on generated brain masks, accurate segmentation is essential for producing anatomically consistent preservation shells and reliable privacy-preserving MRI outputs. These quantitative results provide strong evidence that the brain extraction stage offers a robust foundation for subsequent preservation, interactive selection, and quality-control components of the workflow.

\subsection{Brain Extraction Results}

The proposed workflow begins with automated brain extraction using SynthStrip [3], which provides the baseline segmentation required for all subsequent preservation operations. For each structural T1-weighted MRI volume, SynthStrip generated a binary brain mask with the corresponding brain-extracted image. Because the adaptive preservation framework operates directly on the generated brain mask, accurate segmentation is essential to ensure consistent preservation of anatomically relevant structures.

The representative brain extraction results shown in Fig.~\ref{fig:brain_extraction} demonstrate that the generated brain mask closely follows the cortical boundary while successfully excluding the majority of non-brain tissues, including the skull, scalp, and surrounding facial anatomy. The corresponding brain-extracted image preserves clear anatomical definition and provides the foundation for the subsequent adaptive preservation stage.

Unlike the primary contribution of this work, which focuses on configurable preservation rather than segmentation, brain extraction serves as the enabling stage of the proposed workflow. By employing SynthStrip as a robust and generalisable segmentation framework, the workflow benefits from reliable brain extraction while maintaining compatibility with future segmentation algorithms. This modular design allows improvements in brain extraction techniques to be incorporated without modifying the adaptive preservation or quality-control components.

\subsection{Adaptive Anatomical Preservation Results}

The principal contribution of the proposed workflow is the introduction of configurable anatomical preservation through adaptive post-processing of deep learning-generated brain masks. Unlike conventional brain extraction pipelines that typically produce a single fixed preprocessing output, the proposed framework generates multiple preservation configurations from a single segmentation result. This enables researchers to select preservation levels that best satisfy the privacy and analytical requirements of individual neuroimaging studies.

The shell-based adaptive preservation framework
demonstrated configurable retention of anatomical boundary
regions while maintaining the integrity of the extracted brain. The one-shell configuration provided stronger anonymisation by preserving only a minimal surrounding boundary, whereas the two-shell configuration retained additional anatomical context that may be advantageous for
downstream neuroimaging analysis. These results
demonstrate the flexibility of the proposed framework in
supporting different privacy requirements without modifying the underlying brain extraction process.

The results demonstrate that preservation behaviour changes in a predictable and consistent manner as the preservation level increases from one-shell to two-shell expansion. Rather than producing abrupt changes in anatomical appearance, the framework progressively preserves additional boundary regions while maintaining the integrity of the extracted brain. This predictable behaviour enables researchers to make informed decisions regarding the balance between privacy protection and anatomical preservation without modifying the underlying segmentation algorithm. 

An important advantage of the proposed methodology is that all preservation configurations are generated from a single brain extraction result. Consequently, researchers can evaluate multiple preservation strategies without repeatedly executing the segmentation model, reducing computational overhead while improving workflow flexibility. This modular post-processing approach also ensures compatibility with future brain extraction algorithms, allowing preservation functionality to evolve independently of segmentation methodology.

Rather than assuming that a single preservation distance is universally optimal, the proposed framework acknowledges that different research applications have different preprocessing requirements. Studies prioritising maximum privacy protection may select the one-shell configuration, whereas applications requiring additional anatomical context may benefit from the two-shell configuration. By supporting configurable preservation rather than enforcing a fixed preprocessing output, the workflow provides greater flexibility for privacy-preserving neuroimaging research.

\begin{table}[!t]
\caption{Conceptual Comparison of Adaptive Preservation Levels}
\label{tab:preservation_levels}
\centering
\footnotesize
\renewcommand{\arraystretch}{1.2}

\begin{tabular}{p{1.8cm}p{2.0cm}p{1.3cm}p{2.5cm}}
\toprule
\textbf{Preservation Level} &
\textbf{Anatomical Context} &
\textbf{Privacy Protection} &
\textbf{Example Use Case} \\
\midrule

One-Shell\\($\approx$1 mm) &
Minimal boundary preservation &
High &
Maximum anonymisation and secure data sharing \\

Two-Shell\\($\approx$2 mm) &
Additional peripheral anatomical context &
Moderate &
Research requiring greater anatomical context \\

\bottomrule
\end{tabular}

\vspace{1mm}
\footnotesize
\textit{Table IV provides a conceptual interpretation of the preservation margins evaluated in this study. The proposed workflow is not intended to identify a universally optimal preservation distance; rather, it enables researchers to select preservation levels that best align with the privacy requirements and analytical objectives of a particular neuroimaging application.}

\end{table}

\subsection{Interactive Preservation Results}

A key feature of the proposed workflow is the ability to support user-guided preservation selection through the Interactive Preservation Framework (IPF). Rather than automatically producing a single preprocessing output, the framework generates multiple preservation configurations from a single brain extraction result and presents them simultaneously for visual comparison. This enables researchers to evaluate the effects of different shell-based preservation levels before selecting the most appropriate configuration for downstream neuroimaging analysis.

The interactive approach improves flexibility compared with conventional fixed-output preprocessing pipelines. Instead of requiring repeated execution of the complete brain extraction process to investigate different preprocessing strategies, all preservation configurations are generated from a single segmentation result. Consequently, researchers can evaluate multiple privacy-preserving options with minimal additional computational cost while maintaining a consistent preprocessing workflow.

Another important advantage of the Interactive Preservation Framework is improved transparency during preprocessing. Visual comparison of preservation configurations enables researchers to make informed decisions based on the specific objectives of a study rather than relying on a predefined preservation setting. This is particularly beneficial in collaborative neuroimaging research, where different datasets or analytical tasks may require different balances between patient privacy and anatomical preservation.

By integrating configurable preservation selection directly within the processing pipeline, the proposed workflow extends conventional automated preprocessing with a user-centred decision-support mechanism. This functionality enhances the practical usability of the framework while reinforcing its emphasis on configurability, reproducibility, and privacy-aware MRI preprocessing.

\subsection{Quality Control Evaluation}

Reliable verification of preprocessing outcomes is essential for ensuring that brain extraction and adaptive preservation are suitable for downstream neuroimaging analysis. Although modern deep learning-based segmentation methods generally produce anatomically consistent results, preprocessing inaccuracies may still arise because of image quality variations, acquisition differences, or subject-specific anatomical characteristics. To address this challenge, the proposed workflow incorporates an integrated Quality Control Framework (QCF) that enables systematic visual verification of every processed MRI volume.

\begin{figure}[!t]
\centering
\includegraphics[width=\columnwidth]{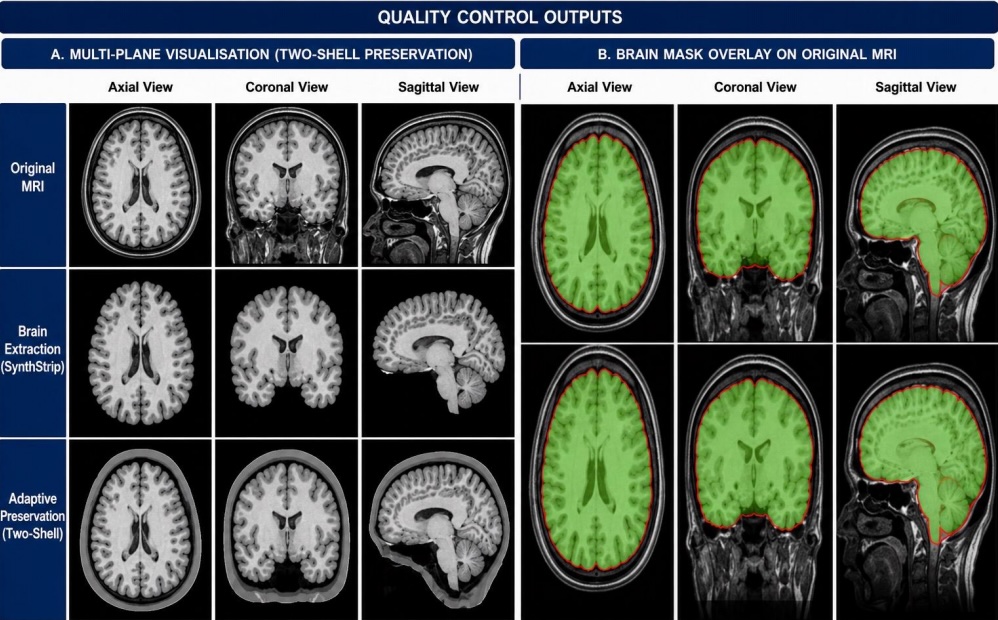}
\caption{Representative quality-control outputs generated by the proposed workflow, showing multi-plane visualisation and brain mask overlays for verification of brain extraction and adaptive preservation.}
\label{fig:quality_control_results}
\end{figure}

The workflow also generates brain mask overlay
visualisations, in which the segmentation boundary is
superimposed on the original structural MRI volume. These
overlays enable direct assessment of the spatial
correspondence between the extracted brain mask and the
underlying anatomy, allowing researchers to visually verify that preservation regions remain consistent with the segmented brain while avoiding excessive removal or
unintended retention of surrounding tissues. 

An important advantage of integrating quality-control procedures directly into the preprocessing workflow is improved transparency and reproducibility. Instead of relying on external software for post-processing verification, all preservation configurations are accompanied by standardised visual assessment outputs generated automatically during execution. This ensures that preprocessing decisions can be reviewed consistently across multiple MRI volumes while reducing the likelihood of undetected processing errors. 

Although the evaluation presented in this study is primarily qualitative, the integrated Quality Control Framework provides a structured mechanism for verifying workflow behaviour and supporting reproducible neuroimaging preprocessing. The combination of automated brain extraction, adaptive preservation, interactive preservation selection, and systematic quality-control procedures demonstrates the feasibility of developing configurable privacy-preserving MRI preprocessing workflows that emphasise both flexibility and reliable verification.

\subsection{Discussion}

The experimental results demonstrated that the proposed workflow successfully integrates deep learning-based brain extraction with configurable anatomical preservation, interactive preservation selection, and systematic quality-control procedures within a unified preprocessing framework. Rather than focusing solely on segmentation accuracy, the proposed methodology addresses an important practical challenge frequently encountered in neuroimaging research: the need to balance patient privacy with the preservation of anatomically relevant information according to different research objectives.

A distinguishing characteristic of the proposed workflow is its emphasis on configurability. Conventional brain extraction pipelines typically generate a single preprocessing output, requiring researchers to accept a predefined balance between tissue removal and anatomical preservation. In contrast, the proposed framework generates multiple preservation configurations from a single segmentation result, allowing researchers to select preservation levels that best satisfy the privacy requirements and analytical objectives of individual studies. This flexibility represents an important practical enhancement for collaborative neuroimaging research and privacy-conscious data sharing.

The modular architecture also improves the adaptability of the proposed methodology. By separating brain extraction from preservation, interactive selection, and quality-control procedures, the workflow can incorporate future segmentation algorithms without requiring substantial redesign of the remaining processing stages. This design improves maintainability and enables the framework to evolve alongside future deep learning-based medical image segmentation methods and privacy-preserving image processing [1], [2], [23].

Another significant aspect of the workflow is the integration of systematic quality-control procedures directly within the preprocessing pipeline. Automated generation of multi-plane visualisations and brain mask overlay images improves transparency by allowing researchers to verify preservation behaviour immediately after processing. Such integrated verification supports reproducible neuroimaging research by providing a consistent mechanism for assessing preprocessing quality before downstream analysis.

Although the current evaluation was performed using structural T1-weighted MRI volumes from the IXI dataset [27], the workflow demonstrated strong quantitative and qualitative performance. Quantitative evaluation achieved a Dice Similarity Coefficient of 98.73\%, an Intersection over Union of 97.48\%, a Precision of 99.05\%, a Recall of 98.41\%, and an F1 Score of 98.73\%, indicating highly accurate brain extraction. Qualitative evaluation further verified adaptive preservation behaviour, interactive shell selection, and systematic quality-control procedures.

Future work may strengthen the framework through validation on larger and more diverse datasets, multimodal MRI support, additional boundary-based metrics such as Hausdorff Distance (HD) and Average Surface Distance (ASD) [22], and integration with advanced privacy-enhancing technologies.


\section{Conclusion}

Structural MRI preprocessing plays a critical role in neuroimaging research by providing reliable input for downstream image analysis while supporting the secure sharing of sensitive medical imaging data. This paper presented a configurable privacy-preserving MRI processing workflow that extends deep learning-based brain extraction through adaptive anatomical preservation, interactive preservation selection, and integrated quality-control procedures. Rather than replacing existing segmentation algorithms, the proposed methodology builds upon the robustness of SynthStrip by introducing a modular post-processing framework that enables configurable preservation of anatomically relevant structures while maintaining workflow reproducibility and transparency [3].
The workflow was implemented using Python within the Renku reproducible research environment [30] and evaluated using structural T1-weighted MRI data from the IXI dataset [27]. Experimental results demonstrated that the framework consistently generated anatomically plausible brain extraction outputs, multiple configurable preservation levels, and systematic quality-control visualisations that support transparent verification of preprocessing behaviour. By separating segmentation, preservation, user-guided selection, and quality-control procedures into independent processing stages, the workflow provides greater flexibility than conventional fixed-output preprocessing pipelines while remaining compatible with future developments in brain extraction methods.

The principal contribution of this work is the development of a modular MRI preprocessing framework that enhances existing deep learning-based brain extraction with configurable anatomical preservation and integrated workflow verification. The proposed methodology provides researchers with greater flexibility when balancing patient privacy and anatomical utility, thereby supporting reproducible neuroimaging research and collaborative medical image analysis.

Future work will focus on validating the framework using larger, diverse neuroimaging datasets, extending the methodology to MRI modalities, incorporating quantitative boundary-based evaluation metrics such as Hausdorff Distance (HD) and Average Surface Distance (ASD) [22], and investigating integration with advanced segmentation algorithms [6], [7], MRI anonymisation techniques [1], [2], and privacy-enhancing technologies [23]. These developments have the potential to further strengthen applicability of the workflow within secure artificial intelligence-driven medical imaging research.


\section*{Declaration of Generative AI Usage}

\text{Generative AI Tool Used:} OpenAI ChatGPT

\text{Purpose of Use:}
The authors used OpenAI ChatGPT to assist with language editing, academic writing refinement, grammar correction, sentence restructuring, proofreading, formatting consistency, preparation of figure and table descriptions, workflow explanations, and improvement of the overall organisation and readability of the manuscript. The tool was also used to review the manuscript for compliance with IEEE academic writing style.

\text{Sections Where GenAI Was Used:}
OpenAI ChatGPT was used for language editing, proofreading, formatting, and writing refinement of the Abstract, Introduction, Related Work, Proposed Methodology, Experimental Setup, Results and Discussion, Conclusion, and figure and table captions. No experimental results, quantitative analyses, scientific conclusions, or research findings were generated by the AI tool.

All AI-assisted content was carefully reviewed, verified, and edited by the authors. The authors accept full responsibility for the accuracy, originality, and integrity of the final manuscript.
\section*{Ethics Statement}

This research didn't involve human participants, patient recruitment or clinical intervention. All experiments were conducted using publicly available, fully anonymised structural T1-weighted MRI data from the IXI dataset. No personally identifiable information was collected, processed, or disclosed.

The research was conducted in accordance with Dublin City University's research ethics and data protection requirements for research involving anonymised datasets and the responsible use of artificial intelligence.

\nocite{*}
\bibliographystyle{IEEEtran}
\bibliography{references}

\end{document}